\documentclass[10pt,letterpaper]{article}

\usepackage[pagenumbers]{wacv} 

\usepackage{graphicx}
\usepackage{amsmath}
\usepackage{amssymb}
\usepackage{booktabs}
\usepackage[linesnumbered,ruled,vlined,procnumbered]{algorithm2e}
\usepackage{multirow}
\usepackage{subcaption}
\usepackage{verbatim}
\usepackage{multicol}
\usepackage{stfloats}
\usepackage{tabularx}
\usepackage{float}
\usepackage{makecell}

\usepackage[pagebackref,breaklinks,colorlinks]{hyperref}

\usepackage[capitalize]{cleveref}
\crefname{section}{Sec.}{Secs.}
\Crefname{section}{Section}{Sections}
\Crefname{table}{Table}{Tables}
\crefname{table}{Tab.}{Tabs.}

\def\wacvPaperID{2648} 
\def\confName{WACV}
\def\confYear{2025}

\begin{document}

\title{Blind Image Deblurring with FFT-ReLU Sparsity Prior}

\author{Abdul Mohaimen Al Radi\textsuperscript{*}, Prothito Shovon Majumder\textsuperscript{*}, Md. Mosaddek Khan\\
University of Dhaka\\
{\tt\small abdulmohaimenal-2018925300@cs.du.ac.bd, prothitoshovon-2018725302@cs.du.ac.bd,}\\
{\tt\small mosaddek@du.ac.bd}\\ \\
{\textsuperscript{*}These authors contributed equally to this work.}
}
\maketitle

\begin{multicols}{2}
\begin{abstract}
   Blind image deblurring is the process of recovering a sharp image from a blurred one without prior knowledge about the blur kernel. It is a small data problem, since the key challenge lies in estimating the unknown degrees of blur from a single image or limited data, instead of learning from large datasets. The solution depends heavily on developing algorithms that effectively model the image degradation process. We introduce a method that leverages a prior which targets the blur kernel to achieve effective deblurring across a wide range of image types. In our extensive empirical analysis, our algorithm achieves results that are competitive with the state-of-the-art blind image deblurring algorithms, and it offers up to two times faster inference, making it a highly efficient solution.\footnote{The source code will be available in the authors' Github profile.}
\end{abstract}


\section{Introduction}
\label{sec:intro}


Blind image deblurring aims to recover a sharp image \( I \) from a blurred image \( B \) without knowing the blur kernel \( k \). Mathematically, the blurred image \( B \) is modeled as:

\[ B = I \otimes k + \eta \]

where \( \otimes \) denotes convolution and \( \eta \) represents noise. The task is to estimate both \( I \) and \( k \) from \( B \). This is formulated as an optimization problem:

\[ \min_{I, k} \| B - I \otimes k \|^2 + \lambda \mathcal{R}(I, k) \]

where \( \| \cdot \|^2 \) measures the difference between the observed and predicted images, \( \mathcal{R}(I, k) \) is a regularization term to enforce constraints, and \( \lambda \) is a regularization parameter. The goal is to recover both \( I \) and \( k \) by balancing data fidelity and regularization.

Blind image deblurring addresses a fundamental challenge in image processing: restoring clarity from blurred images when the exact nature of the blur is unknown. The algorithmic approach generalizes the image deblurring process, helping data-driven methods perform better.

A common approach in blind image deblurring is the maximum a posteriori (MAP) framework, where the latent image and the blur kernel are alternately optimized. Various heuristics involving the latent image, the kernel, or both have been developed in this framework \cite{TFChan, Fergus, KrishnanNormSparse, LevinMarg, PanText, QiShan, XuL0}. This results in new priors that produce effective outcomes. These methods often target specific types of images, as they rely on image properties that aid optimization \cite{PanText}. Additionally, the computation of some priors can be resource-intensive \cite{DCP, Chen}, which can affect the overall efficiency of deblurring techniques. While these approaches demonstrate promising results in their respective domains, their performance can deteriorate when applied to broader image categories, leading to inconsistent outcomes in some cases.

Mao \etal \cite{DeepRFT} show that applying a specific sequence of operations—removing negative frequencies from a blurry image, reconstructing it in the spatial domain, and subtracting half of the original image—yields an implicit blur kernel representation. This process helps in understanding the blur's direction and intensity.
Building on this insight, we discovered that, when applied to a sharp image, this sequence results in a null kernel, characterized by significantly fewer non-zero elements compared to any blur kernel's implicit representation. \par
Our thorough empirical analysis demonstrates that the \(L_0\) norm obtained after applying this sequence of operations to a blurry image consistently exceeds that of its sharp counterpart. Leveraging this property, we incorporated it into our optimization model, leading to a highly efficient algorithm. The linear nature of our calculation allows our algorithm to be up to two times faster than popular blind image deblurring methods, while still delivering competitive results in terms of PSNR, SSIM, and error ratio.\par
\begin{figure*}
\centering
\includegraphics[width=\textwidth ]{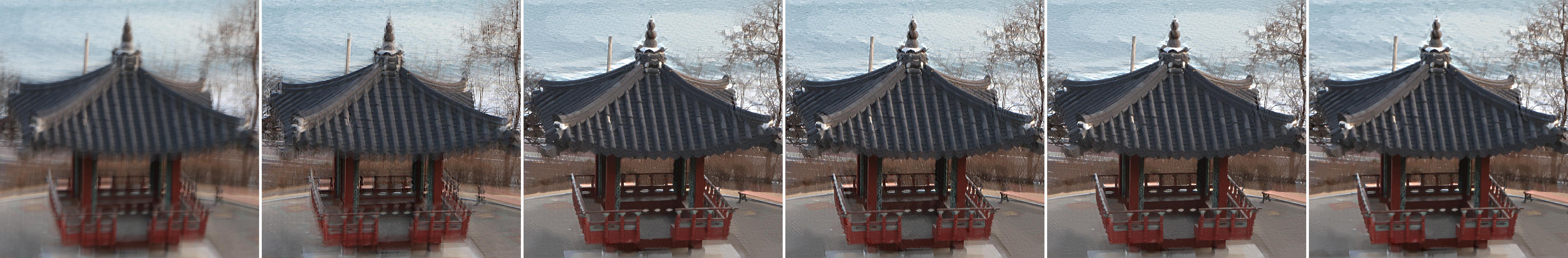}
\caption{Results of our blind image deblurring algorithm, compared with other state-of-the-art algorithms. From left to right: 1) Input blurry image, followed by results from 2) Chen \etal \cite{Chen} 3) Wen \etal \cite{PMP} 4) Pan \etal \cite{PanText}, 5) Pan \etal \cite{DCP} and 6) our algorithm.}
\label{FirstImg}
\end{figure*}

Our main contributions are summarized as follows:
\begin{enumerate}
    \item We propose a new prior, the ReLU Sparsity prior, which allows us to obtain faithful information about the blur kernel regardless of the nature of the image, while simultaneously applying optimization techniques based on the change in sparsity that occurs due to convolution. \cref{FirstImg} shows an example of deblurred output from our algorithm.
    \item Our algorithm achieves competitive performance based on the PSNR, SSIM, and error ratio metrics when compared with the state-of-the-art blind image deblurring algorithms \cite{ChoFast, DCP, PanPhase, QiShan, Xu2010TwoPhaseKE, Fergus, KrishnanNormSparse} while achieving significantly lower inference time on high-resolution images when compared using images from benchmark datasets \cite{KohlerDataset, LevinEval, SunDataset}.
\end{enumerate}

In the following sections, we begin by discussing related works pertinent to our study, followed by a detailed explanation of our proposed prior and the empirical evidence supporting the generalizability of the novel property. We then present our experimental results, starting with quantitative analysis and concluding with qualitative evaluations.


\section{Related Works}
\label{sec:relWork}
Due to the joint estimation of the blur kernel and the latent sharp image from the input blurry image, single-image deblurring is an under-constrained problem, which is typically formulated in MAP frameworks \cite{Caisheng}. Initial advances relied on statistical priors and salient edge detection \cite{ChoFast, Xu2010TwoPhaseKE, XuL0, QiShan, Fergus, KrishnanNormSparse, LevinMarg}, although the existence of strong edges in latent images is not guaranteed.

Different image priors and likelihood estimations have been proposed to improve the efficiency of MAP frameworks. Such regularizations or assumptions about the blur kernel or the latent image include L\textsubscript{0} regularised prior \cite{PanText, Xu2014InverseKF}, dark channel prior \cite{DCP}, extreme channel prior \cite{Extreme}, patch prior \cite{Ljube}, local binary pattern prior \cite{LBP}, latent structure prior \cite{LatStruc}, learned image prior \cite{Lerenhan}, uniform blur \cite{LevinEval, XuL0}, non-uniform blur with multiple homographies \cite{JointDepth, SoftSeg}, local maximum difference prior \cite{LMD}, internal patch recurrence \cite{IPR}, Laplacian prior \cite{Laplus}, tri-segment intensity prior \cite{TSI}, graph-based image prior \cite{Bai2018GraphBasedBI}, constant depth \cite{MDF, Depth}, in-plane rotation \cite{Rotation}, and forward motion \cite{Forward}. Fergus \etal \cite{Fergus} use variational Bayesian inference to learn an image gradient prior. Levin \etal \cite{LevinEval} show that this method can avoid trivial solutions that naive MAP-based methods can possibly be unable to.

Sparsity priors have been shown to be useful in kernel estimation in MAP frameworks. Krishnan \etal \cite{KrishnanNormSparse} employed a normalized sparsity prior in their MAP framework for kernel estimation. Xu and Jia \cite{Xu2010TwoPhaseKE} introduced a two-phase technique for single-image deblurring. First, they estimated the blur kernel using edge selections and ISD optimizations. This was followed by non-blind deconvolution with total variation and an added Gaussian prior \cite{JoshiPSF, ChoFast}.

Some comparatively newer approaches designed for domain-specific deblurring, such as low-illumination \cite{LightStreak}, text \cite{Document, Cho2012TextID, PanText} and face \cite{FaceExemplar} images utilize inherent statistical characteristics of their domains. Chen \etal \cite{Chen} propose the local maximum gradient (LMG) prior, which utilizes the diminishing of the maximum value of a local patch during the blurring process. Wen \etal \cite{PMP} propose the patch-wise minimum pixel (PMP) prior, which is built on the fact that the local minimal pixel would have a higher value in intensity after the blurring process due to the smoothing effect on the image pixels of the process itself. Both these priors use non-linear optimization schemes in MAP frameworks and perform well on both natural and specific images.

\section{FFT-ReLU Sparsity Prior}
In this section, we discuss the sparsity property of applying RFT operation and develop the FFT-ReLU Sparsity prior to formulate an objective function, in order to estimate the latent sharp image and blur kernel.
\subsection{Sparsity and RFT}
\label{sec:SparsityRFT}
\textbf{Sparsity:} To describe our work, we begin by discussing the effect of convolution on the sparsity of images. In standard deblurring models, blurry images are denoted as the convolution of the sharp image and a blur kernel. Essentially, the expression can be written as:
\begin{equation}
    B(x,y)=\sum_{i=0}^{k-1}\sum_{j=0}^{k-1}S(x+i,y+j)\cdot K(i,j)
\end{equation}
Here, \((x,y)\) denotes the pixel location, and the blur kernel \(K\) has a size of \(k \times k\). \(S\) and \(B\) are the sharp image and the blurry image, respectively. \par
Since the output of convolution essentially represents a locally weighted linear combination of the input, it is essentially less sparse (meaning it has more nonzero elements, due to the multiplicative and additive properties of convolution) than the input. Therefore, a blurry image has fewer pixels which are very low in brightness (i.e. it has more nonzero elements), compared to the corresponding sharp image.\par
\textbf{RFT:} As introduced by Mao \etal \cite{DeepRFT}, RFT is defined as a function that computes the following for an image:
\begin{enumerate}
\setlength\itemsep{-0.4em}
    \item Computes the FFT of the input blurry image
    \item Applies the ReLU activation function on it
    \item Computes the inverse FFT
    \item Subtracts half of the image from the result
\end{enumerate}
Formally, we can describe RFT as:
\begin{equation}
    RFT(I) = \mathcal{F}^{-1}(ReLU(\mathcal{F}(I))) - \frac{1}{2}I
\end{equation}
Mao \etal show that faithful information about the blur pattern, such as the direction and the level of blur, can be discovered by applying this sequence of operations.  We observe that applying RFT to a sharp image results in fewer non-zero elements than its blurry counterpart, which provides us with the opportunity to penalize the nonzero results from RFT since those indicate the presence of blur.\par
\begin{figure*}[t]
\centering
    \begin{subfigure}[b]{0.48\textwidth}
        \centering
        \includegraphics[height=6.5cm, width=6.5cm]{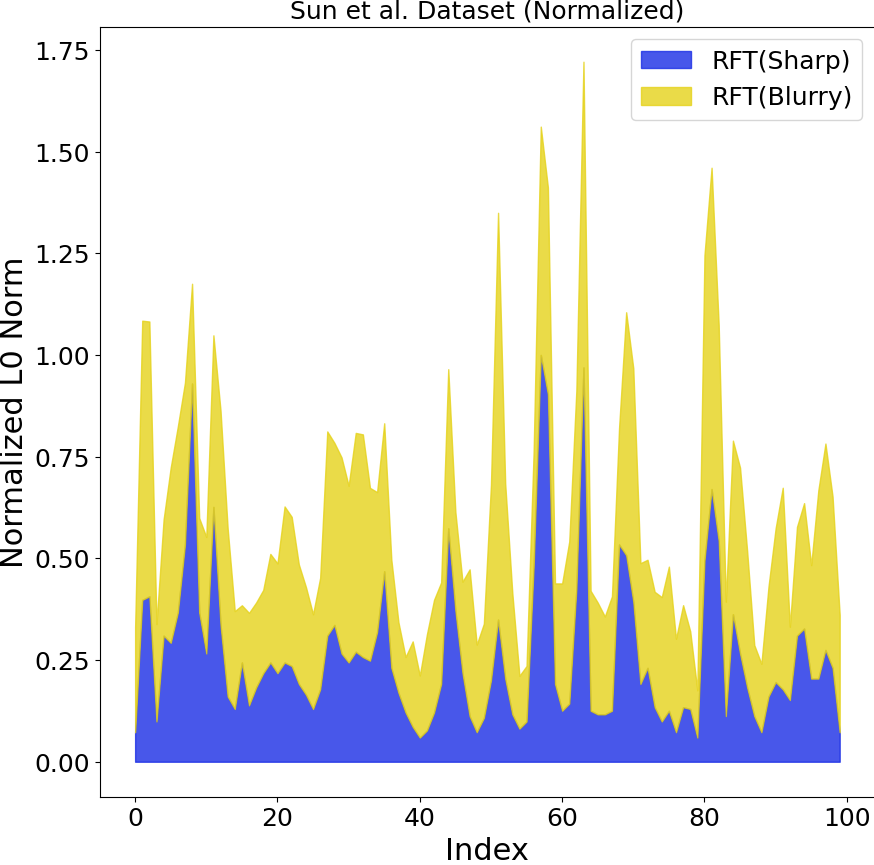}
    \end{subfigure}
    \hfill
    \begin{subfigure}[b]{0.48\textwidth}
        \centering
        \includegraphics[height=6.5cm,width=6.5cm]{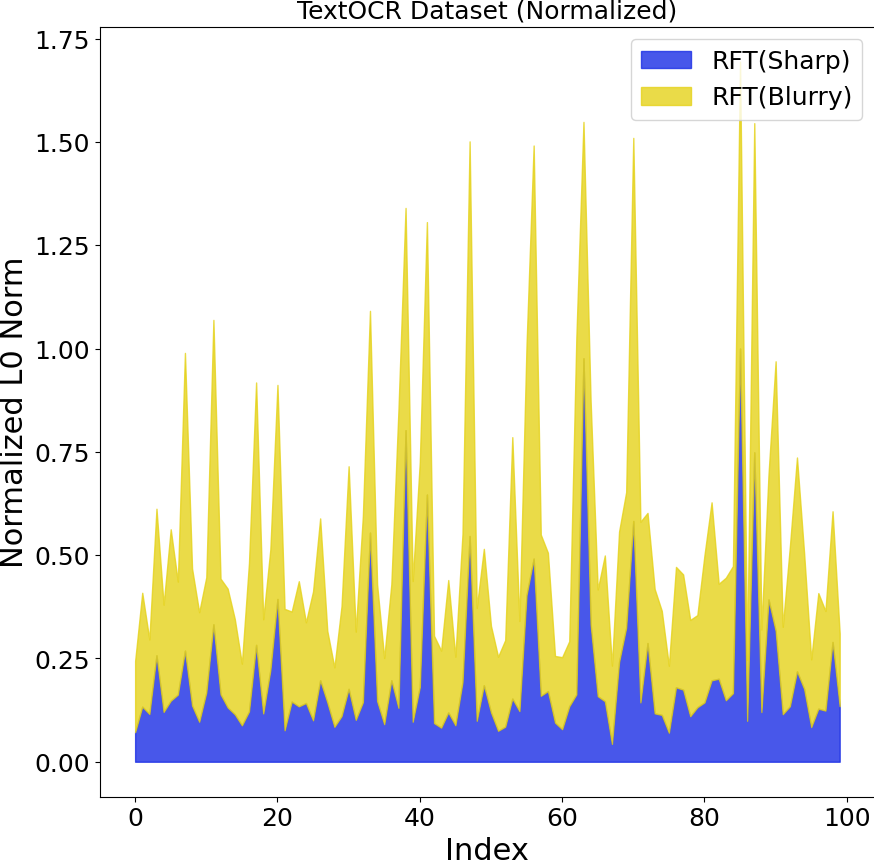}
    \end{subfigure}
    \hfill
    \vspace{1em}
    \begin{subfigure}[b]{0.48\textwidth}
        \centering
        \includegraphics[height=6.5cm,width=6.5cm]{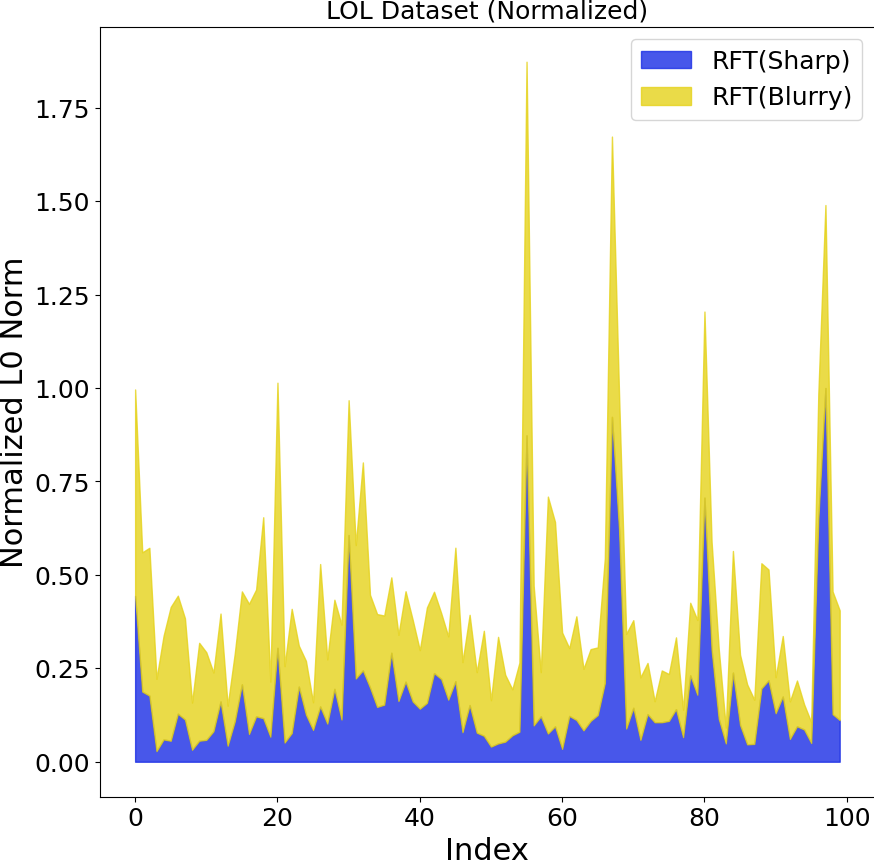}
    \end{subfigure}
    \hfill
    \begin{subfigure}[b]{0.5\textwidth}
        \centering
        \includegraphics[height=6.5cm,width=6.5cm]{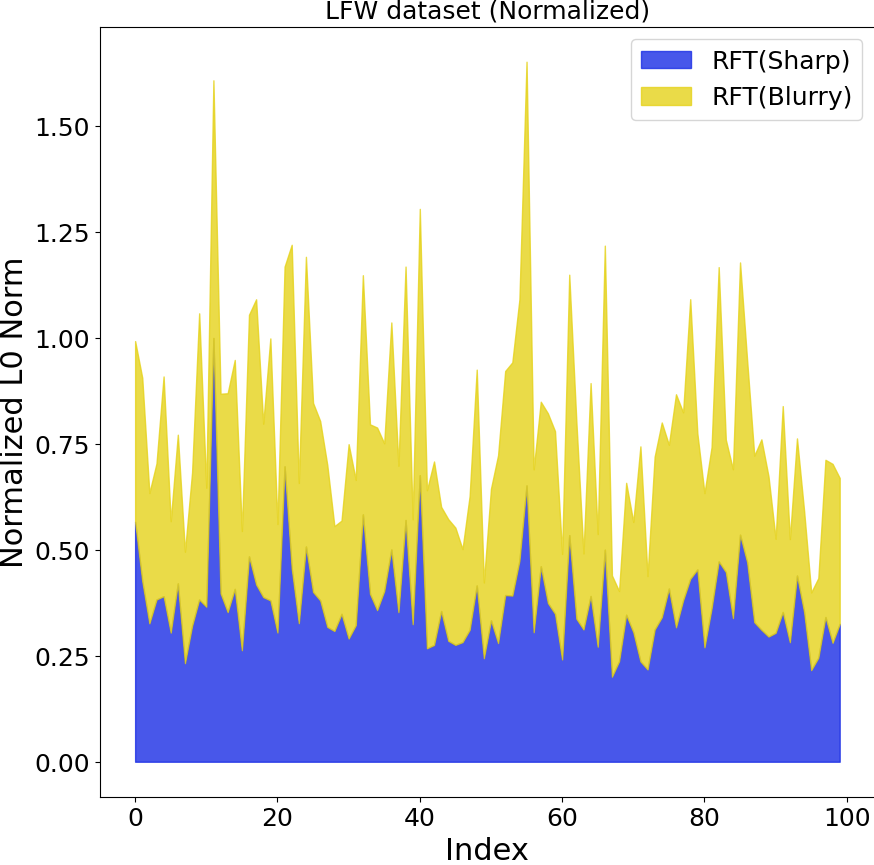}
    \end{subfigure}
    \caption{Decreased Sparsity in Blurry Images after RFT, as demonstrated in images from Sun \etal \cite{SunDataset}, TextOCR dataset, \cite{singh2021textocr} LOL dataset, \cite{loldataset} and LFW dataset \cite{LFWTech}.}
    \label{RFT bars}
\end{figure*}

We apply the RFT operations on the images from various datasets, including one provided by Sun \etal \cite{SunDataset}, datasets of text images \cite{singh2021textocr}, low-light images \cite{loldataset}, face images \cite{LFWTech} and observe the values of the L\textsubscript{0} norm to empirically substantiate our observation in \cref{RFT bars}. Since only Sun \etal provided blurry images with their corresponding sharp pairs, we generated blurry images from the provided sharp images for the other datasets by using blur kernels provided by Chen \etal \cite{ChenNight} randomly. We can see the change in L\textsubscript{0} norm in blurry images from its sharp counterpart when RFT operation is applied to both of them in \cref{RFT bars}.
\subsection{Model and Optimization}
\label{sec:model}
Adding \(RFT(I)\) to standard formulations of image deblurring, we express our objective function as follows:
\begin{equation}
    \operatorname*{min}_{I,k}||I \otimes K - B||^{2}_{2} + \alpha||k||^{2}_{2} + \beta|| \nabla I||_{0} + {\lambda||RFT(I)||_{0}}
    \label{eq:eqfive}
\end{equation}
Here, the first term ensures that the given blurred image and the convolution of the estimated latent image and blur kernels are as similar as possible. The second term applies regularisation on the kernel. The third term retains large gradients and discards smaller ones \cite{PanText, XuL0}. \(\alpha, \beta\) and \(\lambda\) are penalty parameters. From \cref{eq:eqfive}, we solve alternately for the \(I\) and \(k\) using coordinate descent using following equations respectively:
\begin{equation}
    \operatorname*{min}_{I}\|I\otimes k-B\|_{2}^{2}+\beta|| \nabla I||_{0} + \lambda||RFT(I)||_{0}\label{est_I}
\end{equation}
\begin{equation}
\operatorname*{min}_{k}\|I\otimes k-B\|_{2}^{2}+\alpha\|k\|_{2}^{2} \label{est_K}
\end{equation}
Our algorithm can be partitioned into blind deconvolution and non-blind deconvolution segments. In blind deconvolution, the computation of the latent image and the blur kernel is done without an exact PSF by using priors or heuristics. The non-blind deconvolution removes the ringing artifacts that can appear from blind deconvolution.
\subsection{Blind Deconvolution}
\label{bdconv}
The purpose of blind deconvolution is to estimate the sharp image \(I\) and the blur kernel \(k\) from the input image \(B\).

\textbf{Estimating I:}
The non-linear \(RFT\) function and L\textsubscript{0} regularisation make the minimization of \cref{est_I} challenging in terms of computation. Half-quadratic splitting methods \cite{HQS} are used to handle \(L_{0}\) minimization. We split the gradients into \(g = (g_{x},g_{y})\) for gradients across x and y-axis, and introduce auxiliary variable \(h\) for \(RFT(I)\). Rewriting \cref{est_I}, we now have:
\begin{multline}
    \operatorname*{min}_{I,h,g}||I \otimes k - B||^{2}_{2} + \gamma||\nabla I - g||^{2}_{2} +\beta||RFT(I) - h||^{2}_{2} \\ +  \mu||g||_{0} + \lambda||h||_{0}
    \label{est_I_upd}
\end{multline}
where \(\gamma\) and \(\beta\) are penalty parameters. Solutions for individual variables (\(I, u,\) or \(g\)) can be calculated when the others are held constant. In order to use \cref{est_I_upd} in \cref{alg:blind_dconv} to estimate the latent sharp image, we transform this in the frequency domain and reach a closed-form solution in \cref{soln_I}, \cref{soln_h}, and \cref{soln_g}.\par
\begin{figure}[H]
\centering
\includegraphics[width=\columnwidth]{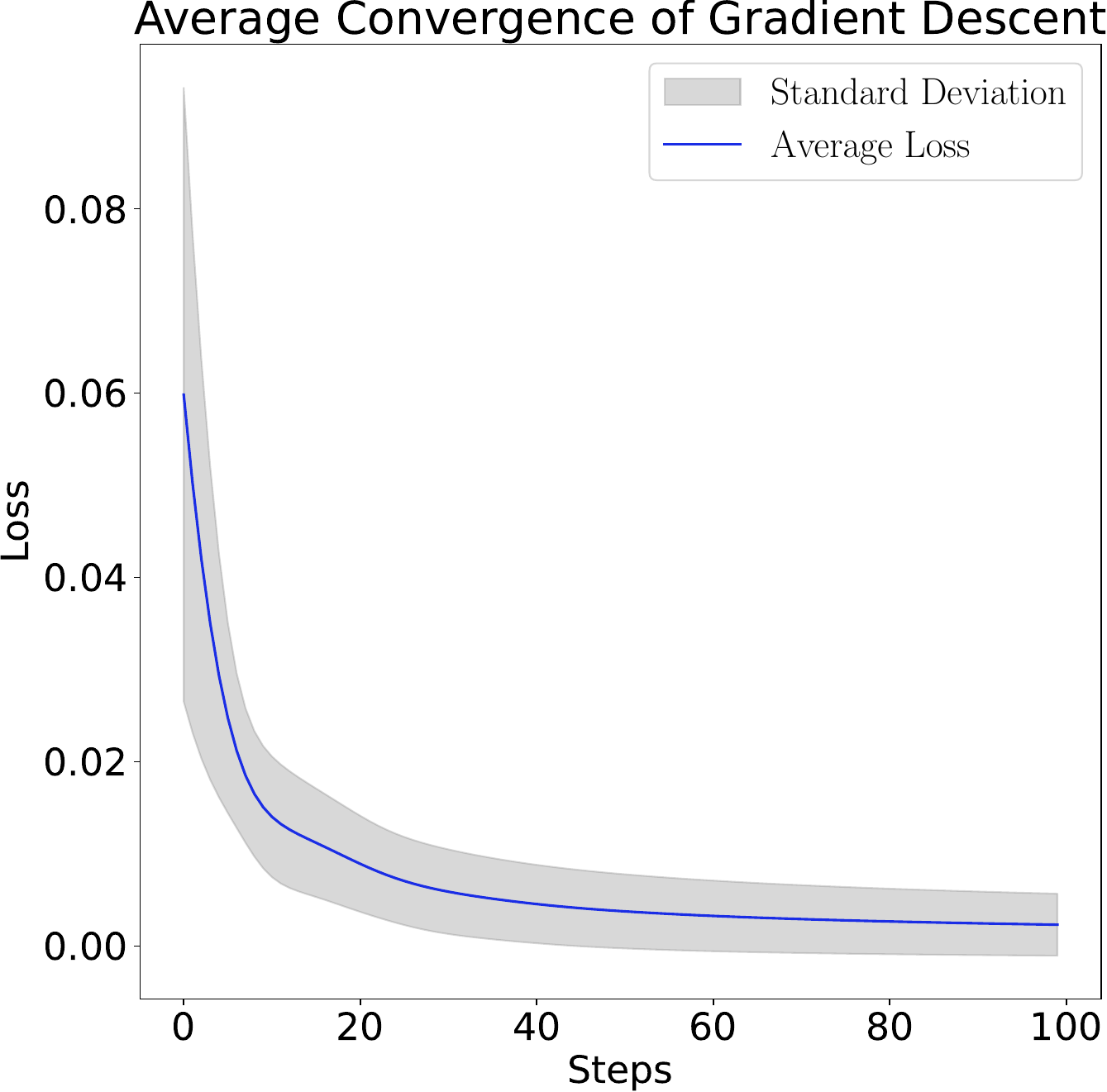}
\caption{Using gradient descent, FI and RFT(I) converge reasonably within 100 steps.}
\label{Grad Desc Std Dev}
\end{figure}

In order to solve for the nonlinear \(RFT(I)\), we observe that a linear operator \(F\) can be used as a substitute when applied to the \(I\) in vector form i.e. for the latent image, \(FI\) = RFT(\(I\)).
To compute the values of \(F\), we use gradient descent with Adam optimizer \cite{AdamOptimizer}. Using repeated reconstruction, we iteratively compute \(F\) from the previous estimation of \(I\).
We look at all images from TextOCR dataset \cite{singh2021textocr}, LOL dataset \cite{loldataset}, and LFW dataset \cite{LFWTech} and observe that \(FI\) closely approximates RFT(\(I\)) and converges reasonably within 100 steps in Figure \cref{Grad Desc Std Dev}.

Given F, we can solve for I from:
\begin{equation}
    \operatorname*{min}_{I}||\textbf{T}_{k}\textbf{I} - \textbf{B}||^{2}_{2} + \gamma||\nabla \textbf{I} - \textbf{g}||^{2}_{2} + \beta||\textbf{FI} - \textbf{h}||^{2}_{2}\label{eq:I_linear}
\end{equation}
where \(\textbf{T}_{k}\) is a Toeplitz matrix of k, which is multiplied with vectors using FFT \cite{Wang_SIAM_2008}. \textbf{B, g} and \textbf{u} are denoted in their vector forms, respectively.
The closed-form solution for \(I\) can be derived as follows:
\begin{equation}
\begin{aligned}
    \textbf{I} &= \frac{\textbf{T}_{k}^{T}\textbf{B} + \gamma\nabla^{T}\textbf{g} + \beta\textbf{F}^{T}\textbf{h}}{\textbf{T}_{k}^{T}\textbf{T}_{k}+\gamma\nabla^{T}\nabla+\beta\textbf{F}^{T}\textbf{F}}\\
    &= \mathcal{F}^{-1}(\frac{\overline{\mathcal{F}(\textbf{K})}\mathcal{F}(\textbf{B}) + \gamma \overline{\mathcal{F}(\nabla)}\mathcal{F}(\textbf{g}) + \beta \overline{\mathcal{F}(\textbf{F})}\mathcal{F}(\textbf{h})}{\overline{\mathcal{F}(\textbf{K})}\mathcal{F}(\textbf{K})+ \gamma \overline{\mathcal{F}(\nabla)}\mathcal{F}(\nabla) + \beta \overline{\mathcal{F}(\textbf{F})}\mathcal{F}(\textbf{F})}) \label{soln_I}
\end{aligned}
\end{equation}
The detailed derivation of this solution is presented in the supplementary material.
Given I, we can solve for \(h\) from:
\begin{equation}
    \operatorname*{min}_{h} \beta||RFT(I) - h||^{2}_{2} + \lambda||h||_{0}
    \label{soln_h}
\end{equation}
and for \(g\) from:
\begin{equation}
    \operatorname*{min}_{g}\gamma||\nabla I - g||^{2}_{2} + \mu||g||_{0}
    \label{soln_g}
\end{equation}
using element-wise minimization. The solution for \(h\) can be written as:
\begin{equation}
  h=\left\{{\begin{array}{r l}{RFT(I),}&{|RFT(I)|^{2}\geq{\frac{\lambda}{\beta}}\colon}\\ {0,}&{{\mathrm{~otherwise}}}\end{array}}\right.
  \label{elem}
\end{equation} 
The solution for \(g\) can also be written in a similar manner, using element-wise minimization techniques.\par

\textbf{Estimating k:}
\cref{est_K} is a least squares problem when \(I\) is given. Due to the proven accuracy of gradient-based methods for estimating kernels in \cite{XuL0,ChoFast,LevinMarg}, \(k\) is computed from:
\begin{equation}
    \operatorname*{min}_{k}||\nabla I \otimes k - \nabla B||^{2}_{2} + \alpha||k||^{2}_{2} \label{soln_k}
\end{equation}
FFTs are used for solving \cref{soln_k}, using techniques mentioned in \cite{XuL0,PanText,ChoFast}. The negative elements of the kernel are set to zero and the obtained kernel is normalized.
\begin{algorithm}[H]
\DontPrintSemicolon
  \SetKwInOut{Input}{Input}
  \SetKwInOut{Output}{Output}
  \SetAlgoLined
  \Input{$B$: Blurry Image}
  \Output{$I$: Latent Image (Intermediate), $k$: Blur Kernel}
  
  \BlankLine
  $k \gets $initialised from coarse resolution \label{k_init}\; 
  $scale \gets $ computed according to $k$ \label{scale_init}
  \BlankLine
  \For{$i = 0$ \KwTo $max\_scale$}{
    $B_{intermediate} \gets $ downsample($B, scale$)\label{down_sample}\;
    \For{$j = 0$ \KwTo $max\_iter$}{
    $I \gets$ solution of Equation \eqref{est_I_upd} \label{update_I}\;
    $k \gets$ solution of Equation \eqref{est_K} \label{update_k}\;  
    $k \gets $ remove\_isolated\_noise($k$)\label{noise_remove}\;
    $k \gets $ adjust\_psf\_center($k$)\label{adjust_psf}\;
  }
  $k \gets $ estimate\_psf($k, scale$)\label{estimate_psf}\;
  $scale \gets scale + 1$ 
  }
  \BlankLine
  \Return{$I, k$}
  
  \caption{Blind Deconvolution}\label{alg:blind_dconv}
\end{algorithm}
This blind deconvolution algorithm estimates the latent image \(I\) and the blur kernel \(k\), which is necessary for obtaining the latent image. Initially, the kernel \(k\) is initialized to the coarsest resolution  (\cref{k_init}). \cref{scale_init} determines the scale according to the current kernel size. The algorithm iteratively increases the scale. For each scale within a loop, the blurry image \(B\) is down-sampled to the current scale in \cref{down_sample}. We iterate a predetermined number of times within this setting to refine the estimation of \(I\) using Equation \eqref{est_I_upd} in \cref{update_I}, and \(k\) using Equation \eqref{est_K} in \cref{update_k}. Isolated noises are removed by deleting connected components within the kernel that do not exceed a specified threshold (\cref{noise_remove}). The kernel is centered after the noise removal in \cref{adjust_psf}. the kernel is then up-sampled with the current scale in \cref{estimate_psf}. All down-sampling and up-sampling are done using bi-linear interpolation.

\subsection{Non-blind Deconvolution}
\begin{algorithm}[H]
\DontPrintSemicolon
  \SetKwInOut{Input}{Input}
  \SetKwInOut{Output}{Output}
  \SetAlgoLined
  \Input{$I$: Intermediate Latent Image, $K$: Intermediate Blur Kernel}
  \Output{$I$: Intermediate Latent Image with Ringing Artifacts Removed}

  \BlankLine
  \For{$i = 0$ \KwTo $num\_channels$}{
    $I_{c} \gets $ laplacian\_prior\_estimation(\(I_{channel_{I}})\)\; \label{line2}
    $I_{1} \gets $ concatenate($I_{1}, I_{c}$)\; \label{line3}
  }
  $I_{2} \gets$  solution for $I$ using equation \cref{est_I}\; \label{line5}
  $\mathrm{diff}  \gets I_{1} - I_{2}$ \;\label{line6}
  $I\_result \gets I_{1} - $ bilateral\_filter($\mathrm{diff}$) \; \label{line7}
  \BlankLine
  \Return{$I\_result$}
  
  \caption{Non-blind Deconvolution}\label{alg:non-blind_dconv}
\end{algorithm}
For non-blind deconvolution, methods based on Laplacian priors \cite{HyperLaplacian} tend to be effective in preserving fine spatial details. We estimate an image \(I_{1}\) with such priors in \cref{line2} and \cref{line3}. Then, we use the L\textsubscript{0} norm of the gradient from our equation \cref{est_I} to estimate another latent image \(I_{2}\) in \cref{line5}, since gradient-based methods have been shown to provide accurate results in suppressing ringing artifacts. Similar to ringing suppression methods discussed in \cite{QiShan}, we compute a difference map between these two estimated images in \cref{line6}. We use bilateral filtering on the computed difference map and subtract the result from \(I_{1}\) in \cref{line7}, thereby smoothing out artifacts and completing non-blind deconvolution.
\section{Experiment}
In this section, we discuss the parameter settings, quantitative results on datasets \cite{KohlerDataset, LevinEval}, and qualitative results on images from different distributions.
\subsection{Empirical Setting}
The parameters \(\lambda, \mu\) and \(\alpha\) are assigned values \(3 \times 10^{-4}, 0.004\) and \(2\), respectively. We compared the convergence of the algorithm for various values of these parameters with respect to kernel similarity (defined as SSIM between the estimated kernel and ground-truth kernel), and a more detailed analysis of this is provided in the supplementary material. The value of $max\_iter$ is set to be 5 as a trade-off between timing and precision. The non-blind deconvolution techniques mentioned by Pan \etal \cite{PanText} are followed. We provide a Python codebase for blind image deblurring on single images. In order to ensure fairness in comparisons with pre-existing algorithms that were not implemented using Python, we avoided using GPU acceleration. However, since the previous works on blind image deblurring were implemented primarily on MATLAB, the translation to Python offers the opportunity for GPU access in the future as well as extension with deep learning methodologies. 
\subsection{Quantitative Results}
In this section, we evaluate our proposed algorithm with the state-of-the-art algorithms based on the metrics of PSNR, SSIM \cite{SSIM}, and error ratio. Datasets from K{\"o}hler \etal \cite{KohlerDataset} and Levin \etal \cite{LevinEval} are widely-used benchmarks, and are used for evaluating our work against other blind image deblurring algorithms.\par
\begin{figure}[H]
    
    \centering
    \begin{subfigure}[b]{0.49\textwidth}
        \centering
        \includegraphics[width=8cm, height = 6.4cm]{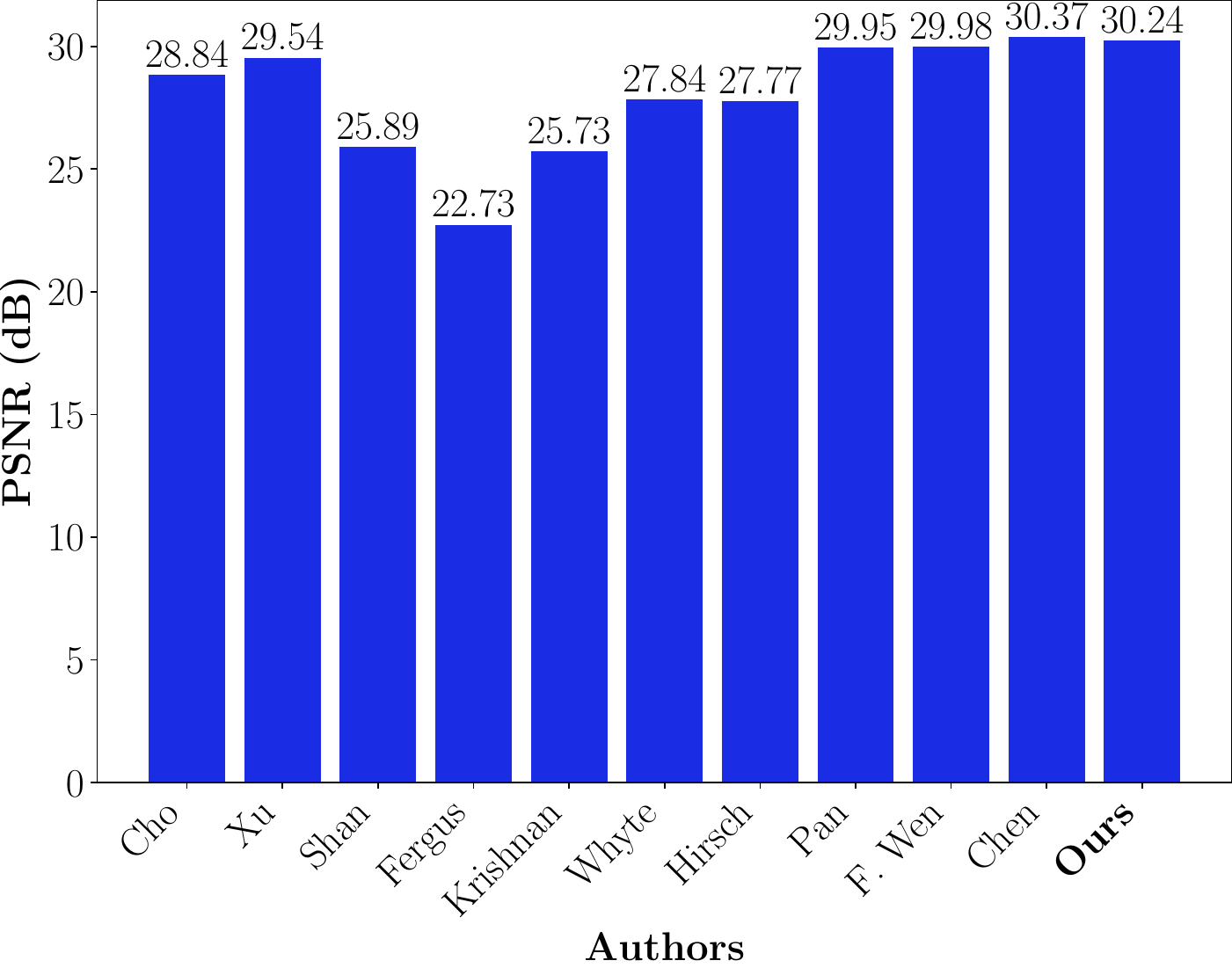}
        \caption{PSNR comparison with other algorithms}
    \label{KohlerPic1}
    \end{subfigure}
    \hfill
    \begin{subfigure}[b]{0.49\textwidth}
        \centering
        \includegraphics[width=8cm, height = 6cm]{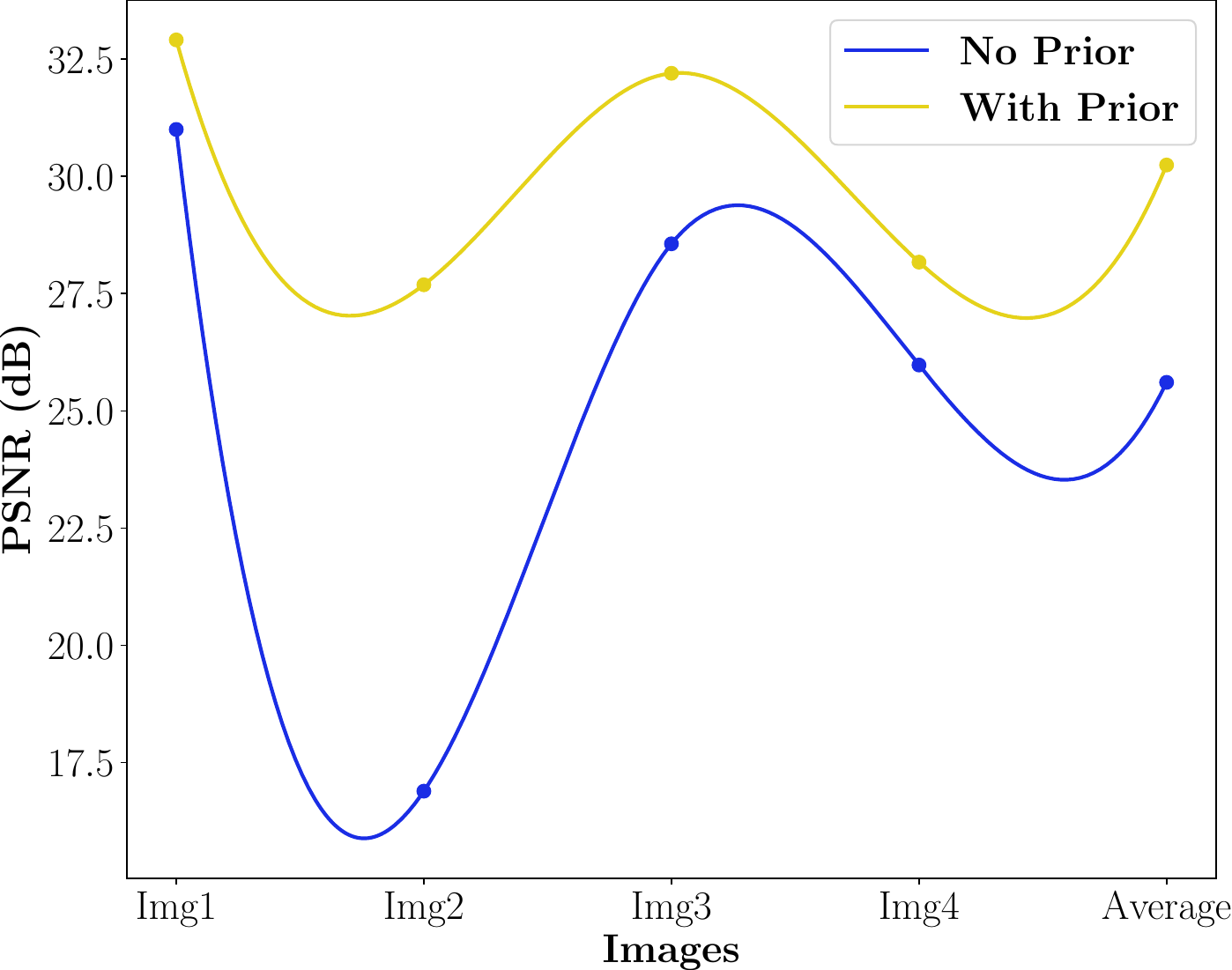}
        \caption{PSNR with vs without our prior}
    \label{KohlerPic2}     
    \end{subfigure}
    
    \caption{PSNR comparison for K{\"o}hler \etal dataset }
\end{figure}
\textbf{PSNR on K{\"o}hler \etal dataset:} There are a total of 32 images. For each of the 4 images and 8 kernels, there are 199 corresponding ground truth images for computing PSNR. We calculated the results of our algorithm for all of these images and plotted the average for each of the 4 images and also for the average of the 4 images against other statistical deblurring algorithms from \cite{QiShan, Fergus, KrishnanNormSparse, XuL0, DCP, ChoFast, Whyte_IJCV_2012, Hirsch_ICCV_2011 } in \cref{KohlerPic1}.
We also presented a comparison of PSNR results when our prior is used versus when it is not used in \cref{KohlerPic2}.\par
\textbf{Levin \etal dataset:} This dataset also contains 4 images and 8 kernels each. The images are grayscale and have dimensions of \(255 \times 255\). We compare our PSNR, SSIM, error ratio, and inference times with the works in \cite{ChoFast, DCP, PanPhase, Chen, PMP}.

\begin{table}[H]
     \captionsetup{justification=centering} 
    \centering
    \scalebox{0.8}{
    \begin{tabular}{c|c|c|c|c}
        \toprule
        \thead{} & \thead{PSNR} & \thead{SSIM} & \thead{Error ratio} &\thead{Average inference \\ time (seconds)}\\
        \midrule
        \cite{DCP} & 27.54 & 0.8626 & 1.2076 & 109.6088 \\
        \cite{PanPhase} & \textbf{28.38} & \textbf{0.925} & \textbf{0.8776} & 15.0949\\
        \cite{PMP} & 26.1235 & 0.83637 & 1.48 & 18.61  \\
        \cite{Chen} & 26.48693 &  0.8515 & 1.01234 & 65.2 \\
        Ours & 28.34 & 0.887 & 1.78 & \textbf{8.057} \\
        \bottomrule
    \end{tabular}
    }
    \caption{Comparison on Levin \etal dataset}
    \label{tab:my_table}
\end{table}
In \cref{tab:my_table}, we show that our method achieves the best inference time while performing competitively in terms of PSNR, SSIM, and inference time.\par
\textbf{Inference Time Comparison:} 
We compare our work with \cite{DCP} and \cite{PanPhase} in this section, using datasets from Levin \etal \cite{LevinEval}, K{\"o}hler \etal \cite{KohlerDataset}  and Sun \etal \cite{SunDataset}, respectively. We report the inference times along with the image dimensions in the table below.\par
\begin{table}[H]
     \captionsetup{justification=centering} 
    \centering
    \begin{tabular}{lccc}
        \toprule
        \multirow{2}{*}{\textbf{Algorithm}} & \multicolumn{3}{c}{\textbf{Inference Time (seconds)}} \\
        \cmidrule(lr){2-4}
        & \textbf{\(255 \times 255\)} & \textbf{\(800 \times 800\)} & \textbf{\(1024 \times 800\)} \\
        \midrule
        Pan \etal\cite{DCP} & 109.609 & 1550.932 & 2280.716	 \\
        Pan \etal\cite{PanPhase} & 15.095 & 210.981 & 269.872	 \\
        Wen \etal\cite{PMP} & 18.61 & 54.416 & 66.134 \\
        Chen \etal\cite {Chen} & 65.2 & 755.43 & 794.186 \\
        Ours & \textbf{8.057} & \textbf{39.5764} & \textbf{55.72}	 \\
        \bottomrule
    \end{tabular}
    \caption{Inference Times of Algorithms on Images of Different Dimensions}
    \label{tab:inference_times}
\end{table}
Our approach achieves significant speed gains: nearly two times lower compared to  \cite{PanPhase}, as presented in \cref{tab:inference_times}.\par
\subsection{Visual Results}
In this section, we present the results of our algorithm on blurry images from different domains, which we compare with other state-of-the-art blind image deblurring algorithms. The images are taken from datasets provided by Lai \etal \cite{LaiDataset}, Levin \etal \cite{LevinEval} and K{\"o}hler \etal \cite{KohlerDataset}.
\end{multicols}

\begin{figure*}
\label{visualresult1}
\centering

\includegraphics[width=\columnwidth ]{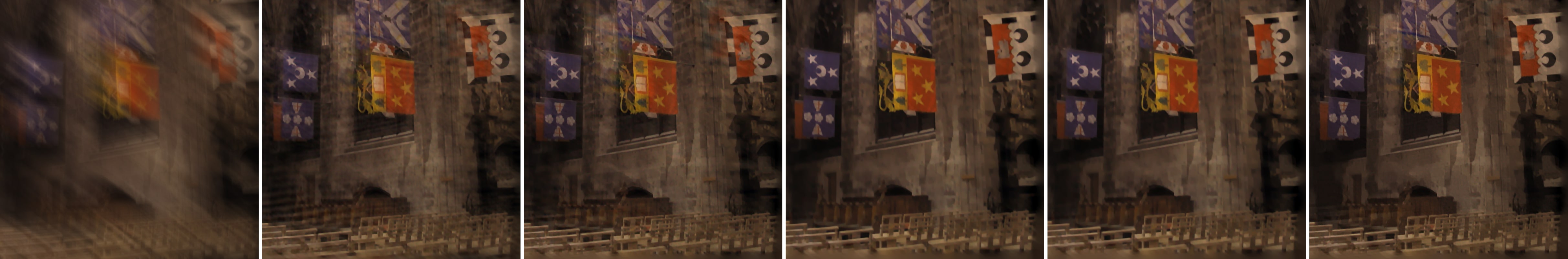}
\includegraphics[width=\columnwidth ]{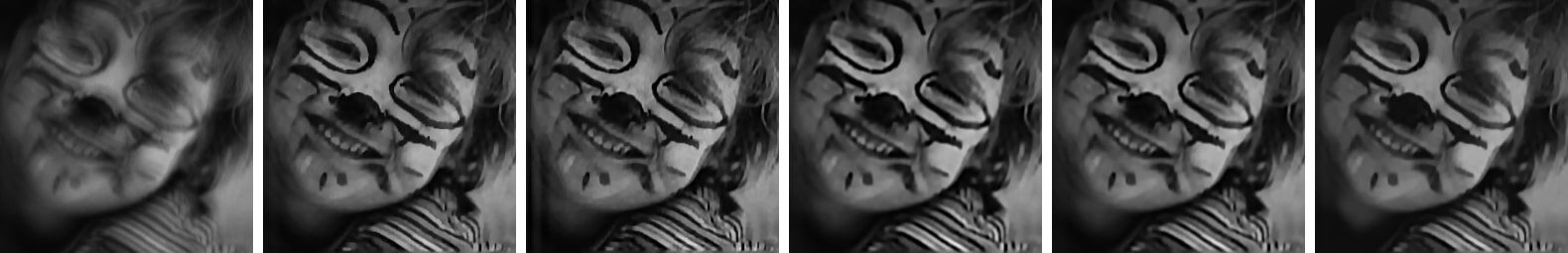}
\includegraphics[width=\columnwidth ]{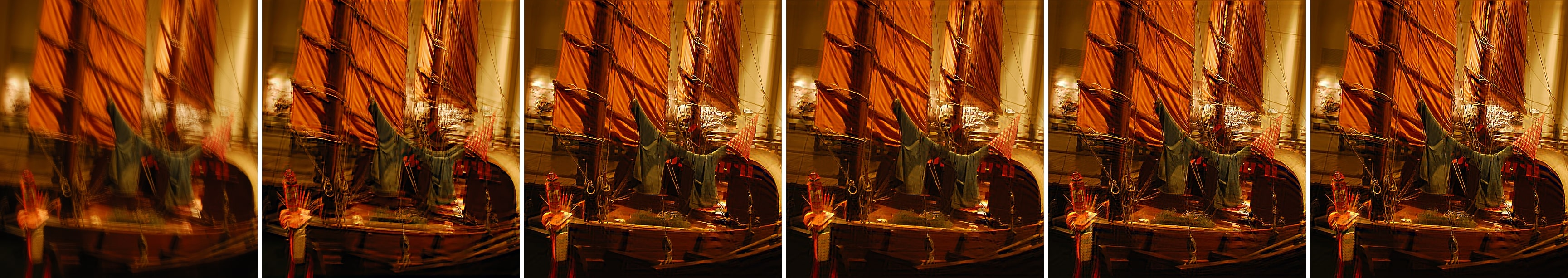}
\includegraphics[width=\columnwidth, height = 3cm ]{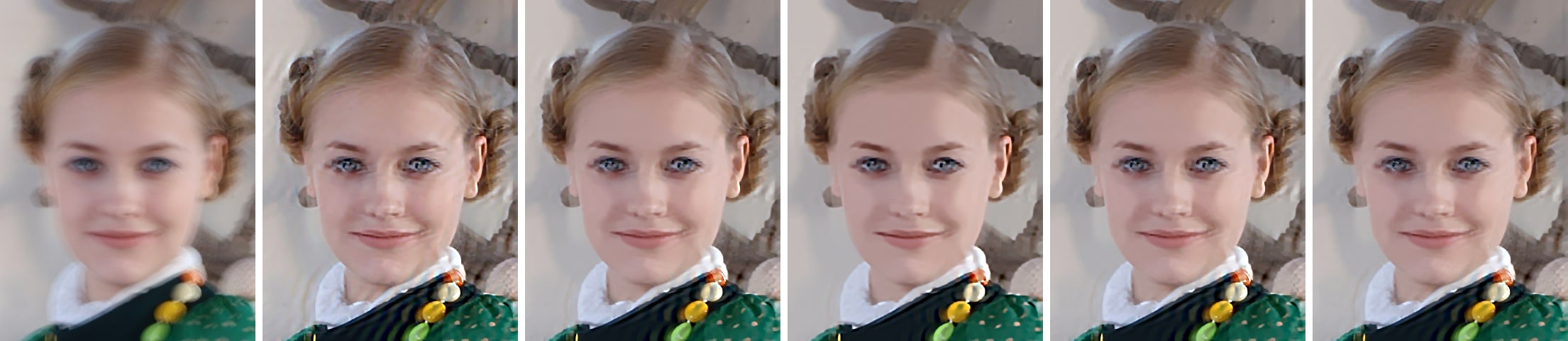}
\includegraphics[width=\columnwidth ]{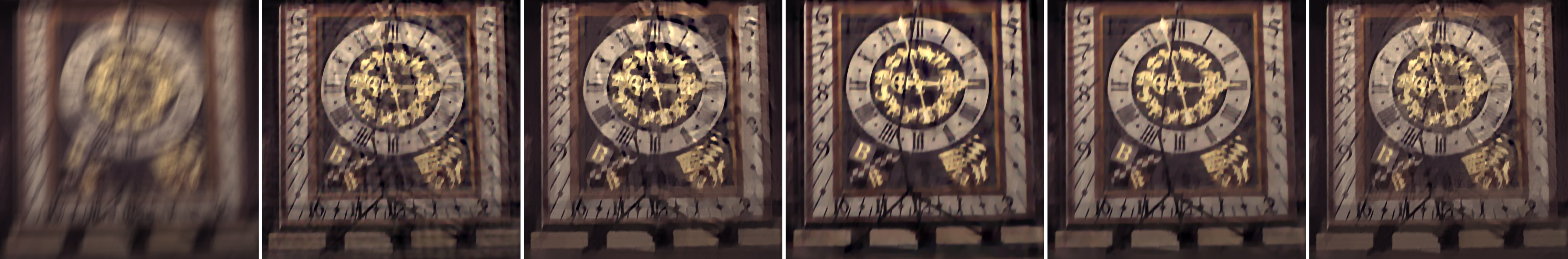}
\includegraphics[width=\columnwidth]{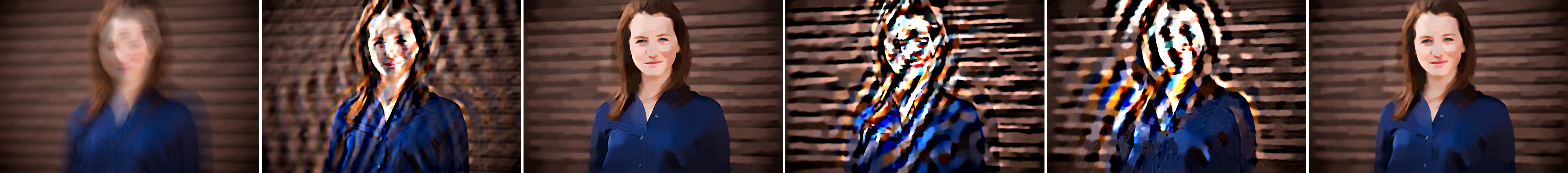}
\includegraphics[width=\columnwidth]{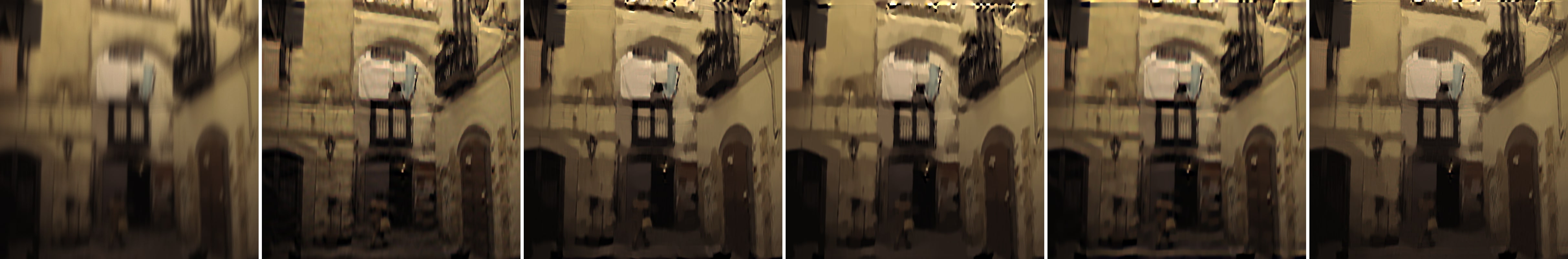}
\caption{Results of our blind image deblurring algorithm, compared with other state-of-the-art algorithms. From left to right: 1) Input blurry image, followed by results from 2) Chen \etal \cite{Chen} 3) Wen \etal \cite{PMP} 4) Pan \etal \cite{PanText}, 5) Pan \etal \cite{DCP} and 6) our algorithm.}
\label{rezaalt}
\end{figure*}
\cleardoublepage
\begin{figure*}
\includegraphics[width=\columnwidth ]{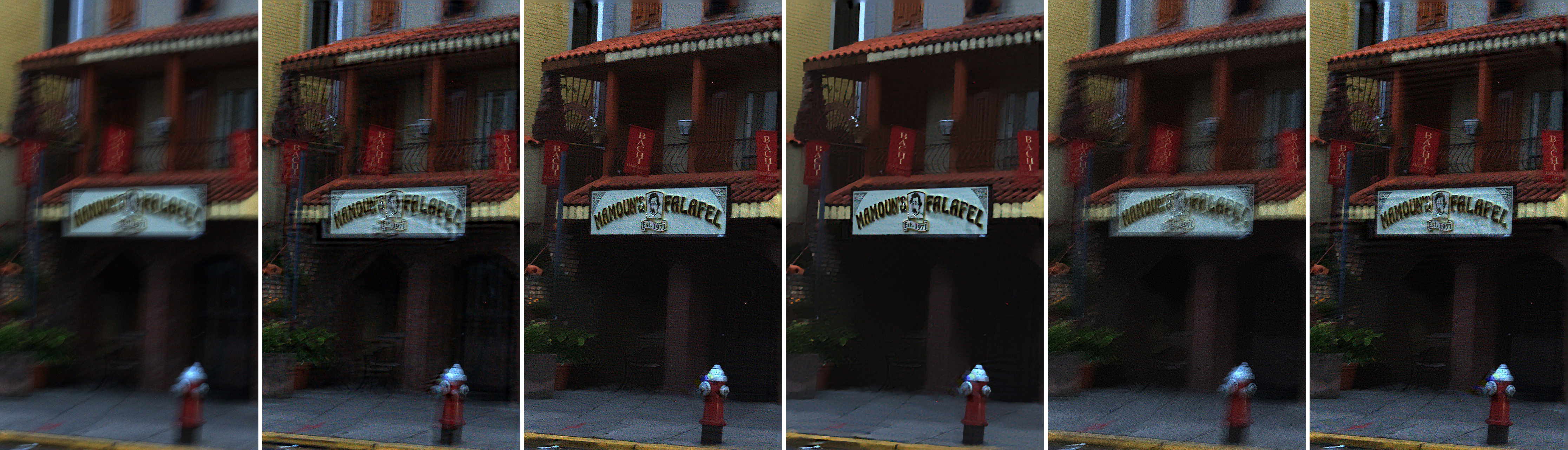}   
\includegraphics[width=\columnwidth ]{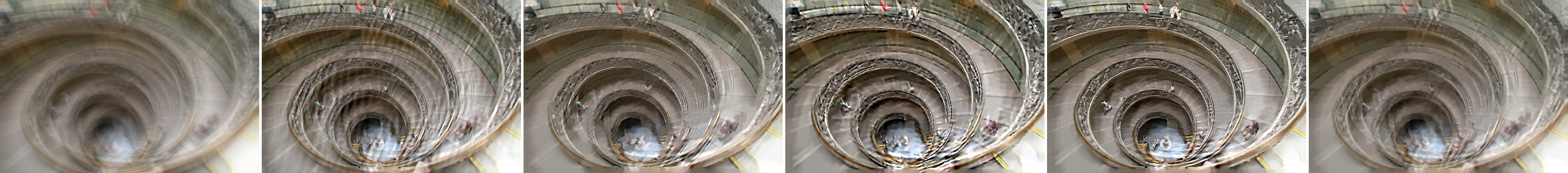}
\includegraphics[width=\columnwidth ]{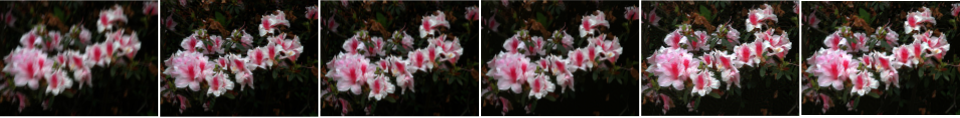}

\includegraphics[width=\columnwidth ]{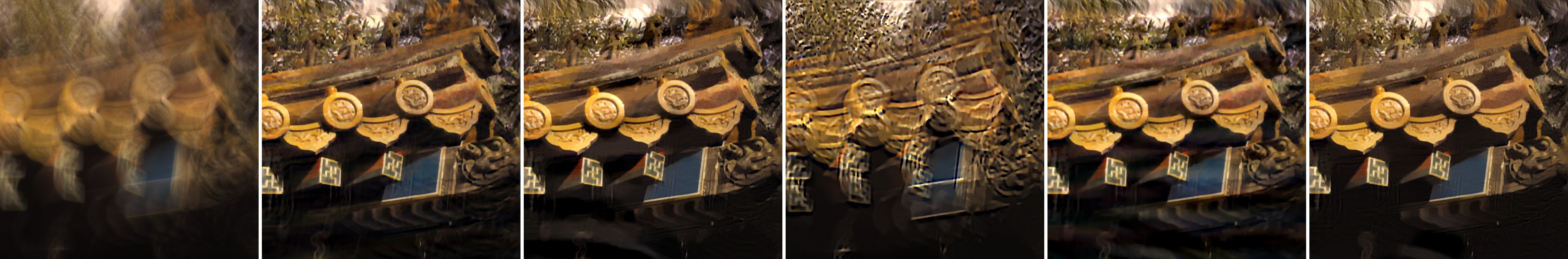}

\label{viz2}
\caption{More results of our blind image deblurring algorithm, compared with other state-of-the-art algorithms. From left to right: 1) Input blurry image, followed by results from 2) Chen \etal \cite{Chen} 3) Wen \etal \cite{PMP} 4) Pan \etal \cite{PanText}, 5) Pan \etal \cite{DCP} and 6) our algorithm.}
\end{figure*}

\begin{multicols}{2}
We observe that our algorithm provides competitive results in terms of visual appeal, as well as qualitative measures mentioned in the previous section, for images irrespective of the distribution or domain they are taken from.
\section{Conclusions and Remarks}

\label{sec:conclusion}
In this paper, we propose a new prior for blind image deblurring based on observations made on the effects of convolution on the sparsity of an image. We made use of Fourier transforms in order to obtain the frequency components in order to separate the negative phase(s), which enabled us to obtain the blur kernel implicitly through information on blur level and blur direction. The ReLU Sparsity prior can be used to penalize a higher L\textsubscript{0} norm, which is an indicator of the blurriness of an image. For the non-convex optimization problem, we use half-quadratic splitting strategies. Our algorithm provides competitive performance on qualitative metrics such as PSNR, SSIM, and error ratio while providing up to two times lower inference times than the established state-of-the-art blind deblurring algorithms. \par
 \textbf{Future work}: We plan to explore the integration of our algorithm's rapid inference and portable framework with deep learning architectures to further enhance deblurring evaluations. Additionally, we aim to develop alternative mathematical strategies to reduce the computational demands associated with gradient descent in RFT image computation. Another area of focus will be leveraging inherent image attributes such as saturation, intensity, and illumination to optimize the estimation of multiple kernels, thereby improving the recovery of the latent image without compromising processing speed.
{\small
\bibliographystyle{ieee_fullname}
\bibliography{egbib}
}
\end{multicols}

\end{document}